\documentclass[10pt,twocolumn,letterpaper]{article}

\usepackage[pagenumbers]{cvpr} 

\makeatletter
\@namedef{ver@everyshi.sty}{}
\makeatother

\usepackage{graphicx}
\usepackage{amsmath}
\usepackage{amssymb}
\usepackage{booktabs}
\usepackage{multirow}
\usepackage{pgfplots}
\usepackage{colortbl}
\pgfplotsset{compat=newest,compat/show suggested version=false}
\newcommand{\blue}[1]{\textcolor{blue}{#1}}
\usepackage{makecell}
\usepackage{pifont}

\usepackage[pagebackref,breaklinks,colorlinks]{hyperref}

\usepackage[capitalize]{cleveref}
\crefname{section}{Sec.}{Secs.}
\Crefname{section}{Section}{Sections}
\Crefname{table}{Table}{Tables}
\crefname{table}{Tab.}{Tabs.}

\begin{document}

\title{PointCMP: Contrastive Mask Prediction for Self-supervised Learning \\ on Point Cloud Videos}

\author{
Zhiqiang Shen$^{1,2}$\footnotemark[1]\ , \ 
Xiaoxiao Sheng$^{1}$\footnotemark[1]\ , \ 
Longguang Wang$^{3}$\footnotemark[2]\ , \ 
Yulan Guo$^{4}$, \ 
Qiong Liu$^{2}$, \ 
Xi Zhou$^{1,2}$\\
$^1$Shanghai Jiao Tong University\ \ \ \ \ \ \ \ \ \ \ 
$^2$CloudWalk\\
$^3$Aviation University of Air Force\ \ \ \ \ \ \ \ \ \ \ 
$^4$Sun Yat-sen University\\
{\tt\small \{shenzhiqiang, shengxiaoxiao\}@sjtu.edu.cn, wanglongguang15@nudt.edu.cn}
}
\maketitle

\renewcommand{\thefootnote}{\fnsymbol{footnote}}
\footnotetext[1]{These authors contributed equally.}
\footnotetext[2]{Corresponding author.}

\begin{abstract}
Self-supervised learning can extract representations of good quality from solely unlabeled data, which is appealing for point cloud videos due to their high labelling cost. In this paper, we propose a contrastive mask prediction (PointCMP) framework for self-supervised learning on point cloud videos. Specifically, our PointCMP employs a two-branch structure to achieve simultaneous learning of both local and global spatio-temporal information. On top of this two-branch structure, a mutual similarity based augmentation module is developed to synthesize hard samples at the feature level. By masking dominant tokens and erasing principal channels, we generate hard samples to facilitate learning representations with better discrimination and generalization performance. Extensive experiments show that our PointCMP achieves the state-of-the-art performance on benchmark datasets and outperforms existing full-supervised counterparts. Transfer learning results demonstrate the superiority of the learned representations across different datasets and tasks.
\end{abstract}

\section{Introduction}
\label{sec:Introduction}
Recently, LiDARs have become increasingly popular in numerous real-world applications to perceive 3D environments, such as autonomous vehicles and robots. 
Point clouds acquired by LiDARs can provide rich geometric information and facilitate the machine to achieve 3D perception. Early works focus on parsing the real world from static point clouds \cite{Chen_2022_CVPR, hu2021learning, zhang2022not}, while recent researches pay more attention to understanding point cloud videos \cite{fan2022point, PST2, wen2022point, fan2021deep}. 
Since annotating point clouds is highly time and labor consuming \cite{xie2020pointcontrast, afham2022crosspoint}, learning from point cloud videos in a self-supervised manner draws increasing interest. 
Although contrastive learning and mask prediction paradigms \cite{chen2020simple, he2020momentum, grill2020bootstrap, zbontar2021barlow, xie2022simmim, he2022masked} have shown the effectiveness of self-supervised learning on images or static point clouds, these methods cannot be directly extended to point cloud videos due to the following three challenges:

    \begin{figure}[t]
    	\centering
    	\setlength{\abovecaptionskip}{0.1cm}
    	\setlength{\belowcaptionskip}{-0.2cm}
    	\includegraphics[width=1\linewidth]{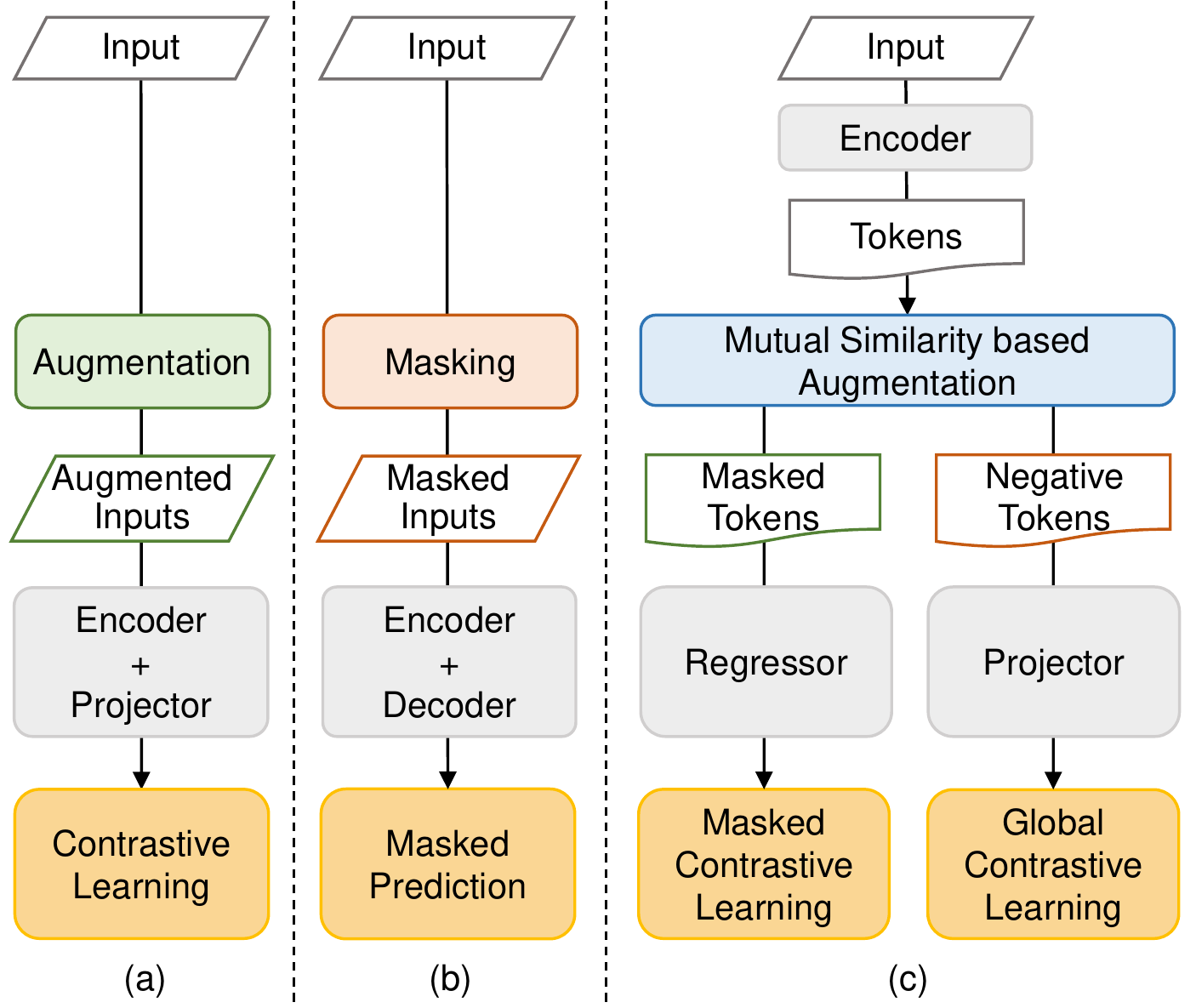}
    	\caption{{A comparison between (a) contrastive learning paradigm, (b) mask prediction paradigm, and (c) our method.}}
    	\label{fig1}
    \end{figure}
    
\textbf{(i) Multiple-Granularity Information Matters.} The contrastive learning paradigm \cite{chen2020simple, he2020momentum, caron2020unsupervised, caron2021emerging, grill2020bootstrap, zbontar2021barlow} usually focuses on extracting global semantic information based on instance-level augmentations. In contrast, the mask prediction paradigm \cite{dosovitskiy2020image, bao2021beit, xie2022simmim, he2022masked, pmlr-v119-henaff20a} pays more attention to modeling local structures while ignoring global semantics. 
However, since fine-grained understanding of point cloud videos requires not only local spatio-temporal features but also global dynamics \cite{fan2021deep, wen2022point}, existing paradigms cannot be directly adopted.


\textbf{(ii) Sample Generation.} The contrastive learning paradigm is conducted by pulling positive samples while pushing negative ones \cite{wang2022importance, chen2020simple, he2020momentum, caron2020unsupervised, grill2020bootstrap, chen2021exploring, zbontar2021barlow}, and the mask prediction paradigm learns representations by modeling the visible parts to infer the masked ones \cite{bao2021beit, xie2022simmim, he2022masked, wang2022bevt, yu2022point, pang2022masked}. 
Both paradigms rely heavily on the augmented samples at the input level.
Further, as demonstrated in several works \cite{kalantidis2020hard, ge2021robust, DBLP, AttentionMask}, self-supervised learning can significantly benefit from proper hard samples. However, the spatial disorder, temporal misalignment, and uneven information density distribution impose huge challenges on hard sample generation for point cloud videos at the input level. 



\textbf{(iii) Leakage of Location Information.} The mask prediction paradigm usually learns to reconstruct masked raw signals by modeling visible ones \cite{bao2021beit, xie2022simmim, he2022masked, wang2022bevt, yu2022point, pang2022masked}. 
For images, the contents are decoupled from the spatial position such that positional encoding is provided as cues to predict masked regions. However, for point clouds with only $xyz$-coordinates, positional encoding may be used as shortcuts to infer the masked points without capturing geometric information \cite{pang2022masked, liu2022masked}.

In this paper, we propose a contrastive mask prediction framework for self-supervised learning on point cloud videos, termed as PointCMP. 
To address challenge (i), our PointCMP integrates the learning of both local and global spatio-temporal features into a unified two-branch structure, and simultaneously conducts self-supervised learning at different granularities (Fig.~\ref{fig1}(c)). 
For challenge (ii), we introduce a mutual similarity based augmentation module to generate hard masked samples and negative samples at the feature level.
To handle challenge (iii), instead of directly regressing the coordinates of masked points, token-level contrastive learning is conducted between the predicted tokens and their target embeddings to mitigate information leakage.

Our contributions are summarized as follows:

\begin{itemize}
\item We develop a unified self-supervised learning framework for point cloud videos, namely PointCMP. Our PointCMP integrates the learning of multiple-granularity spatio-temporal features into a unified framework using parallel local and global branches.
\item We propose a mutual similarity based augmentation module to generate hard masked samples and negative samples by masking dominant tokens and principal channels. These feature-level augmented samples facilitate better exploitation of local and global information in a point cloud video. 
\item Extensive experiments and ablation studies on several benchmark datasets demonstrate the efficacy of our PointCMP on point cloud video understanding.
\end{itemize}



\section{Related Work}
\label{sec:Related Work}
In this section, we first briefly review two mainstream self-supervised learning frameworks. Then, we present recent advances for point cloud video understanding.

\subsection{Contrastive Learning}
Contrastive learning has greatly promoted the development of self-supervised learning \cite{chen2020simple, chen2020big, he2020momentum, caron2020unsupervised, caron2021emerging, grill2020bootstrap, chen2021exploring, zbontar2021barlow, Wang_2021_CVPR}. 
Usually, semantically consistent sample pairs are separately encoded by an asymmetric siamese network, and then contrastive loss aligns them to facilitate the encoder to learn discriminative representations \cite{wang2020understanding, tao2022exploring, wang2022importance}. 
For contrastive learning on images, data augmentation has been widely investigated to generate positive and negative samples to improve the discriminability of representations~\cite{chen2020simple, peng2022crafting,ge2021robust, eccv2022automix}. 

Recently, contrastive learning has also been studied on static point clouds. Specifically, 
Xie~\etal~\cite{xie2020pointcontrast} used random geometric transformations to generate two views of a point cloud and associated matched point pairs in these two views using contrastive loss. Zhang \etal \cite{zhang2021self} constructed two augmented versions of a point cloud and used their global features to setup an instance discrimination task for pre-training.

\subsection{Mask Prediction} 
Mask prediction has demonstrated its effectiveness in numerous computer vision tasks and draws increasing interest \cite{dosovitskiy2020image, bao2021beit, xie2022simmim, he2022masked, wang2022bevt, pmlr-v119-henaff20a}. 
Bao~\etal~\cite{bao2021beit} proposed a BERT-style framework \cite{kenton2019bert} to predict token identities of masked patches based on visible ones. 
Then, Zhou~\etal~\cite{zhou2021image} developed an online tokenizer for better image BERT pre-training. Later, Feichtenhofer \etal \cite{feichtenhofer2022masked} and Tong \etal \cite{tong2022videomae} introduced mask prediction to videos and obtained representations rich in local details by inferring masked spatio-temporal tubes.

Recently, several efforts have been made to extend the mask prediction paradigm to point clouds. Specifically, Yu~\etal~\cite{yu2022point} proposed PointBERT and introduced a masked point modeling task for point cloud pre-training. Pang~\etal \cite{pang2022masked} proposed Point-MAE to reconstruct masked point coordinates using high-level latent features learned from unmasked ones.
Liu~\etal~\cite{liu2022masked} proposed to use binary point classification as a pretext task for point cloud masked autoencoding. 

Most existing contrastive learning and mask prediction methods rely on input-level augmentation to conduct self-supervised learning on static point clouds. Nevertheless, it is intractable to directly extend these methods to point cloud videos as more complicated augmentation operations are required to cover the additional temporal dimension.
To remedy this, we propose to synthesize samples at the feature level based on mutual similarities, which enables reasonable sample generation without considering the unstructured data formats of point cloud videos.

\begin{figure*}[ht]
	\centering
    \setlength{\abovecaptionskip}{0.1cm}
    \setlength{\belowcaptionskip}{-0.3cm}
	\includegraphics[width=1\linewidth]{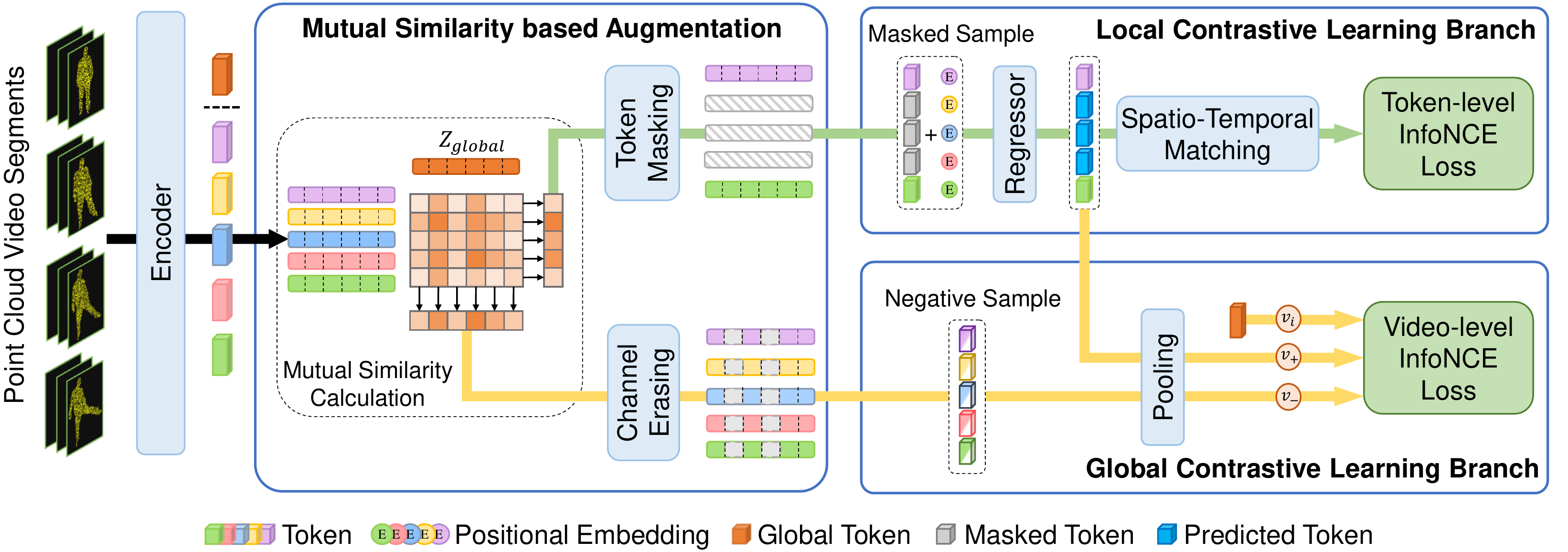}
	\caption{{An overview of our PointCMP.} 
	}
	\label{fig2}
\end{figure*}

\subsection{Point Cloud Video Understanding}
Spatial disorder and temporal misalignment make point cloud videos more challenging to be parsed using a neural network than structured data like images. To leverage the advanced techniques developed for structured data, previous methods transform point clouds into a sequence of bird’s eye views \cite{faf}, voxels \cite{action4d, minke}, and pillar grids \cite{yin}. However, these transformations inevitably lead to the loss of geometric details. Recently, more attention has been paid to learning directly on raw points using 
attention-based models~\cite{p4d, fan2022point, PST2, wen2022point}, convolution-based models~\cite{MeteorNet, pstnet, fan2021deep}, and hand-crafted temporal descriptors~\cite{3dv, zhong2022no}. 
Specifically, Fan~\etal \cite{pstnet} proposed a spatio-temporal decoupled encoder, which alternately performs spatial and temporal convolution to model raw point sequences hierarchically. Then, they further developed P4Transformer~\cite{p4d} that utilizes point spatio-temporal tubes to aggregate local neighborhoods into tokens. 

Despite the huge success of self-supervised learning methods in video understanding \cite{chen2021rspnet, wu2021greedy, han2020self, qian2021spatiotemporal, videomoco, wang2021enhancing, sun2021composable, liang2022self, zhang2022hierarchically}, self-supervised point cloud video understanding is still under-investigated. Recently, Wang~\etal~\cite{wang2021self} designed a pretext task, {namely recurrent order prediction (ROP),} to predict the temporal order of shuffled point cloud segments for self-supervised learning. However, this method can only capture clip-level temporal structures and cannot exploit finer spatio-temporal details. 
To parse a point cloud video, it is important for a self-supervised method to capture both spatio-temporal local structures and global semantics. 
To this end, we develop a unified PointCMP framework that can enable networks to simultaneously learn information with different granularities.

\section{Method}
The architecture of our PointCMP is illustrated in Fig.~\ref{fig2}. Given a point cloud video, it is first uniformly divided into $L$ segments. 
Then, these segments are fed to an encoder to produce tokens $\boldsymbol{Z}\in \mathbb{R}^{L\times N \times C}$ by aggregating local spatio-temporal information, where $N$ means the token number aggregated from each segment and $C$ is the number of channels. 
Meanwhile, a global token $\boldsymbol{Z}_{global}\in \mathbb{R}^{C}$ with global semantics is also obtained following \cite{pstnet}.
Next, $\boldsymbol{Z}_{global}$ and $\boldsymbol{Z}$ are passed to a mutual similarity based augmentation module for online sample generation. Afterwards, a local contrastive learning branch and a global contrastive learning branch are employed to capture multi-granularity information. 

\subsection{Mutual Similarity based Augmentation}
Hard samples have been demonstrated to be critical to the performance of self-supervised learning \cite{kalantidis2020hard, ge2021robust, DBLP}. However, it is challenging to generate hard samples for orderless and unstructured point cloud videos at the input level. To address this issue, we introduce a mutual similarity based augmentation module to synthesize hard samples at the feature level.

\textbf{Hard Masked Samples.}
Our intuition is that reconstruction is easier when tokens sharing higher similarities with the global token are visible. 
Therefore, we are motivated to mask these tokens to synthesize hard masked samples. Specifically, the similarity $\boldsymbol{s}^i$ between the $i$-th token $\boldsymbol{z}^i$ and the global token $\boldsymbol{Z}_{global}$ is calculated as:   
\begin{equation}
\centering
	\boldsymbol{s}^i=\frac{\boldsymbol{z}^i}{\Vert\boldsymbol{z}^i\Vert_{2}}\cdot\frac{\boldsymbol{Z}_{global}}{\Vert\boldsymbol{Z}_{global}\Vert_{2}}.
\end{equation}
Then, the top 40\% tokens with the highest similarities are selected as dominant ones.
Note that, point patches corresponding to adjacent tokens usually share overlapped regions \cite{yu2022point}. That is, token-level masking may introduce shortcuts for mask prediction. To remedy this, segment-level masking is adopted as different segments are isolated.
Specifically, $L_m$ segments with the most dominant tokens are selected with all tokens ($\mathbb{R}^{L_m\times N \times C}$) being masked.
By masking these tokens that share high similarity with the global token, the difficulty of mask prediction is largely increased. It is demonstrated in Sec.~\ref{Sec4.6} that our hard masked samples can facilitate the encoder to achieve much higher accuracy.


\textbf{Hard Negative Samples.}
Our major motivation is that different channels contain information of various importance, and the channels with higher correlation with the global token are more discriminative. Consequently, we synthesize hard negative samples by erasing these channels. 
Specifically, the correlation of the $c$-th channel in the $i$-th token $\boldsymbol{s}_{c}^i$ is calculated as:
\begin{equation}
    \centering
    	\boldsymbol{s}_{c}^i=\frac{\boldsymbol{z}_{c}^i}{\Vert\boldsymbol{z}^i\Vert_{2}}\cdot\frac{\boldsymbol{z}_{c}^{global}}{\Vert\boldsymbol{Z}_{global}\Vert_{2}},
    \label{eq2}
\end{equation}
where $\boldsymbol{z}_{c}^{global}$ is the $c$-th channel of the global token. Then, we rank $\boldsymbol{s}_{c}^i$ to obtain the order of each channel $\boldsymbol{o}_{c}^i$, and sum up the resultant ranks across all tokens:
\begin{equation}
    \centering
    	\boldsymbol{A}_{c}=\sum_{i=1}^{L\times N}{\boldsymbol{o}_c^i},
    \label{eq3}
\end{equation}
Next, the top 20\% channels are selected as principal channels and erased to produce hard negative samples.



\subsection{Local Contrastive Learning Branch} 
\begin{figure}[t]
	\centering
	\setlength{\abovecaptionskip}{0.1cm}
    \setlength{\belowcaptionskip}{-0.3cm}
	\includegraphics[width=0.9\linewidth]{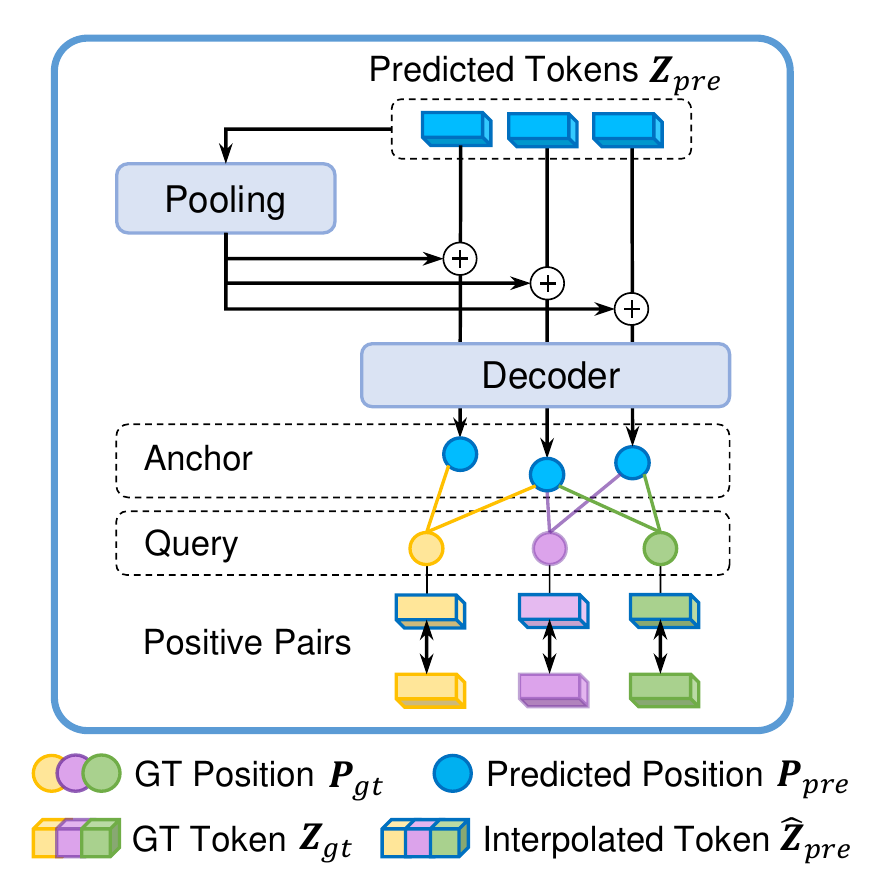}
	\caption{{Network Architecture of our spatio-temporal matching module.}}
	\label{fig3}
\end{figure}

In the local branch, we first generate positional embedding for each token by feeding its spatio-temporal coordinate $(x,y,z,t)$ to a linear layer. Then, these positional embeddings are summed with their tokens and fed to a regressor to predict masked tokens using the context and position cues. 
Next, the predicted tokens $\boldsymbol{Z}_{pre}\in \mathbb{R}^{L_m\times N \times C}$ are passed to a spatio-temporal matching module, as shown in Fig~\ref{fig3}. 
Specifically, $\boldsymbol{Z}_{pre}$ is pooled to obtain a global representation $ \mathbb{R}^{L_m\times C}$, which is then added to $\boldsymbol{Z}_{pre}$. Afterwards, the resultant token is fed into a decoder to predict their position $\boldsymbol{P}_{pre}\in \mathbb{R}^{{L_m} \times {N} \times 3}$. 
Here, a three-layer Transformer \cite{p4d} is adopted as the regressor and FoldingNet \cite{yang2018foldingnet} is used as the decoder. 



As discussed in Sec.~\ref{sec:Introduction}, the positional embeddings may lead to leakage of location information when inferring the coordinates for masked points. To remedy this, we adopt a contrastive loss to associate the representations of predicted tokens $\boldsymbol{Z}_{pre}$ and corresponding groundtruth tokens $\boldsymbol{Z}_{gt}$ {learned by the encoder.} 
Specifically, the tokens located at $\boldsymbol{P}_{gt}$ are obtained through trilinear interpolation by querying $\boldsymbol{Z}_{pre}$ located at $\boldsymbol{P}_{pre}$, resulting in $\boldsymbol{\hat{Z}}_{pre}$.
For the $i$-th token ${\boldsymbol{z}_i}\in\boldsymbol{\hat{Z}}_{pre}$, the corresponding token in $\boldsymbol{Z}_{gt}$ is adopted as the positive sample $\boldsymbol{z}_+$. Meanwhile, other tokens are regarded as negative samples. 
This avoids directly using the token position correspondence to construct sample pairs.
The InfoNCE loss \cite{oord2018representation} is used for training:
\begin{equation}
    \centering
        \mathcal{L}_{\boldsymbol{z}_i}\!=\!-{\log{\frac{\exp{(\boldsymbol{z}_i^T \boldsymbol{z}_+/\tau)}}{{\exp{(\boldsymbol{z}_i^T \boldsymbol{z}_+/\tau)}}+\sum_{\boldsymbol{z}_j\in{\Phi}}{\exp{(\boldsymbol{z}_i^T \boldsymbol{z}_j/\tau)}}}}},
    \label{eq5}
\end{equation}
where $\tau$ is a temperature parameter and $\Phi$ is a negative sample set. 
Through token-level contrastive learning, the encoder can alleviate the shortcuts of positional encoding to capture fine-grained local information.

\subsection{Global Contrastive Learning Branch}
In the global branch, we focus on learning discriminative representations at the video level. 
We take the global token $\boldsymbol{Z}_{global}$ as the query $\boldsymbol{v}_i$, and the resultant tokens produced by the regressor are pooled to obtain the positive sample $\boldsymbol{v}_+$, as shown in Fig.~\ref{fig2}. Meanwhile, the hard negative sample $\boldsymbol{v}_-$ in addition with samples from other videos in batch $\mathcal{B}$ are passed to a max-pooling layer, resulting in negative samples. Then, all samples are projected into the latent space and the InfoNCE loss \cite{oord2018representation} is adopted for training:
\begin{equation}
	\centering
        \mathcal{L}_{\boldsymbol{v}_i}\!=\!-{\log{\frac{\exp{(\boldsymbol{v}_i^T \boldsymbol{v}_+/\tau)}}{{\exp{(\boldsymbol{v}_i^T \boldsymbol{v}_-/\tau)}}\!+\!\sum_{\boldsymbol{v}_j\in{\{\mathcal{B}\cup \boldsymbol{v}_+\}}}^{i\neq j}{\exp{(\boldsymbol{v}_i^T \boldsymbol{v}_j/\tau)}}}}}.
    \label{eq6}
\end{equation}

Overall, the total loss of our PointCMP is defined as:
\begin{equation}
	\centering
        \mathcal{L}_{total}=\mathcal{L}_{NCE}^{local}+\mathcal{L}_{NCE}^{global},
    \label{eq7}
\end{equation}
where $\mathcal{L}_{NCE}^{local}$ refers to the InfoNCE loss in the local branch (Eq. \ref{eq5}), and $\mathcal{L}_{NCE}^{global}$ refers to the InfoNCE loss in the global branch (Eq. \ref{eq6}). With both loss terms, our PointCMP can simultaneously learn both local and global information.

\begin{table*}[ht]
    \centering
	\setlength{\abovecaptionskip}{0.1cm}
    \setlength{\belowcaptionskip}{-0.4cm}
    \caption{Action recognition accuracy (\%) on MSRAction-3D.}
    \setlength{\tabcolsep}{3.8mm}
    \begin{tabular}{l|l|ccccc}
    \toprule
    \multicolumn{2}{c|}{\multirow{2}{*}{\textbf{Methods}}} & \multicolumn{5}{c}{\textbf{\#Frames}}\\
    \cmidrule(r){3-7}
    \multicolumn{2}{l|}{} & 4     & 8     & 12    & 16     & 24 \\
    \midrule
    \multirow{8}{*}{{Supervised Learning}} &
    MeteorNet~\cite{MeteorNet} & 78.11 & 81.14 & 86.53 & 88.21 & 88.50 \\
    & Kinet~\cite{zhong2022no}   & 79.80 & 83.84 & 88.53 & 91.92 & 93.27 \\
    & PST$^2$~\cite{PST2}        & 81.14 & 86.53 & 88.55 & 89.22 & -     \\
    & PPTr~\cite{wen2022point}   & 80.97 & 84.02 & 89.89 & 90.31 & 92.33 \\
    & P4Transformer~\cite{p4d}   & 80.13 & 83.17 & 87.54 & 89.56 & 90.94 \\
    & PST-Transformer~\cite{fan2022point}  & 81.14 & 83.97 & 88.15 & 91.98 & 93.73 \\
    & \cellcolor{gray!20}PSTNet~\cite{pstnet} & \cellcolor{gray!20}81.14 & \cellcolor{gray!20}83.50 & \cellcolor{gray!20}87.88 & \cellcolor{gray!20}89.90 & \cellcolor{gray!20}91.20 \\
    & PSTNet++~\cite{fan2021deep} & 81.53 & 83.50 & 88.15 & 90.24 & 92.68 \\    
    \midrule
    End-to-end Fine-tuning \ \ \ & \textbf{PSTNet + PointCMP} & \textbf{84.02} & \textbf{89.56} & \textbf{91.58} & \textbf{92.26} & \textbf{93.27} \\
    Linear Probing & \textbf{PSTNet + PointCMP} \ \ \ & 78.11 & 88.55 & 90.24 & 91.92 & 92.93 \\
    \bottomrule
    \end{tabular}
    \label{MSRAction-3D}
\end{table*}

\section{Experiments}
In this section, we first present the datasets and implementation details used in the experiments. Then, we compare our PointCMP to previous methods under four widely used protocols, including end-to-end fine-tuning, linear probing, semi-supervised learning, and transfer learning. Finally, we conduct ablation studies to demonstrate the effectiveness of our method.

\subsection{Datasets and Implementation Details}

\noindent\textbf{Datasets.}
We conduct experiments on 3D action recognition and 3D gesture recognition tasks. 
Four benchmark datasets are employed, including NTU-RGBD~\cite{ntu60}, MSRAction-3D~\cite{msr}, NvGesture~\cite{molchanov2016online}, and SHREC'17~\cite{de2017shrec}. 

\begin{itemize}
    \item \textbf{NTU-RGBD}. The NTU-RGBD dataset consists of 56,880 videos with 60 action categories performed by 40 subjects. Following the cross-subject setting of \cite{ntu60}, this dataset is split into 40,320 training videos and 16,560 test videos. 

    \item \textbf{MSRAction-3D}. The MSRAction-3D dataset contains 567 videos with 23k frames. It consists of 20 fine-grained action categories performed by 10 subjects. Following \cite{pstnet}, this dataset is split into 270 training videos and 297 test videos. 

    \item \textbf{NvGesture}. The NvGesture dataset is comprised of 1532 videos with 25 gesture classes. Following \cite{min2020efficient}, this dataset is split into 1050 training videos and 482 test videos. 

    \item \textbf{SHREC’17}. The SHREC’17 dataset consists of 2800 videos in 28 gestures. Following \cite{de2017shrec}, 1960 videos are used as the training set and 840 videos are adopted as the test data.
\end{itemize}

\noindent\textbf{Pre-training Details.}
During pre-training, 16 frames were sampled as a clip from each point cloud video, with 1024 points being selected for each frame. The frame sampling stride was set to 2 and 1 on NTU-RGBD and MSRAction-3D, respectively. Then, we divided each clip into 4 segments and random scaling was utilized for data augmentation. Our model was pre-trained for 200 epochs with a batch size of 80, and linear warmup was utilized for the first 5 epochs. The initial learning rate was set to 0.0003 with a cosine decay strategy. 
The spatial search radius was initially set to 0.5/0.1 on NTU-RGBD/MSRAction-3D and the number of neighbors for the ball query was set to 9. The temperature parameter $\tau$ was set to 0.01/0.1 in the local/global InfoNCE loss term.

\subsection{End-to-end Fine-tuning}
\label{Sec4.2}
We first evaluate our representations by fine-tuning the pre-trained encoder with a linear classifier in a supervised manner. The MSRAction-3D dataset was used for both pre-training and fine-tuning. During fine-tuning, 2048 points were selected for each frame and the pre-trained model was trained for 35 epochs with a batch size of 24. The initial learning rate was set to 0.015 with a cosine decay strategy. Following~\cite{pstnet}, the initial spatial search radius was set to 0.5 and the number of neighbors for the ball query was set to 9. Quantitative results are presented in Table~\ref{MSRAction-3D}.

As we can see, our PointCMP introduces significant accuracy improvements over the baseline trained in a fully supervised manner. Especially, the accuracy achieved using 8/12 frames is improved from 83.50\%/87.88\% to 89.56\%/91.58\%. 
{This shows that our PointCMP can learn beneficial knowledge from point cloud videos in a self-supervised manner, which contributes to higher accuracy after fine-tuning.}

\subsection{Linear Probing}
We then conduct experiments to validate the effectiveness of our PointCMP via linear probing. {The MSRAction-3D dataset was used for both pre-training and linear probing.} Specifically, the pre-trained encoder is frozen and an additional linear classifier is added for supervised training. The experimental settings are the same as Sec.~\ref{Sec4.2}. 

From Table~\ref{MSRAction-3D}, we can see that the pre-trained encoder using PointCMP outperforms the fully supervised baseline even under the linear probing setting. Our method surpasses the baseline under most frame settings with notable margins (e.g., 88.55\%/90.24\% vs. 83.50\%/87.88\% under 8/12 frames). This clearly demonstrates the high quality of the representations learned by PointCMP.

\subsection{Semi-supervised Learning}
We also conduct experiments to evaluate our PointCMP under the setting of semi-supervised learning. 
The cross-subject training set of NTU-RGBD was used for pre-training. Specifically, we used only 50\% training set of NTU-RGBD to fine-tune the pre-trained encoder in a supervised manner. 
Following~\cite{pstnet}, the initial spatial search radius was set to 0.1, the number of neighbors for the ball query was set to 9, and 2048 points were samples for each frame. The model was fine-tuned for 50 epochs with a batch size of 24. The initial learning rate was set to 0.015 with a cosine decay strategy. 

Table~\ref{NTU} compares the quantitative results produced by our PointCMP and previous fully supervised approaches. 
Averaged accuracy over 3 experiments is reported for our method.
It can be observed that our PointCMP achieves comparable performance to the fully supervised baseline even with only 50\% data (88.5\% vs. 90.5\%). 
{This further demonstrates the superiority of the representations learned by our PointCMP.}

\begin{table}[t]
    \centering
    \setlength{\abovecaptionskip}{0.1cm}
    \setlength{\belowcaptionskip}{-0.3cm}
    \small
    \caption{Action recognition accuracy on NTU-RGBD under cross-subject setting.}
    \setlength{\tabcolsep}{1.7mm}
    \begin{tabular}{lc}
    \toprule
    \textbf{Methods}    & \textbf{Accuracy (\%)} \\
    \midrule
    Kinet~\cite{zhong2022no}                 & 92.3 \\
    P4Transformer~\cite{p4d}                 & 90.2 \\
    PST-Transformer~\cite{fan2022point}      & 91.0 \\
    \rowcolor{gray!20}
    PSTNet~\cite{pstnet}                     & 90.5 \\
    PSTNet++~\cite{fan2021deep}              & 91.4 \\
    \midrule
    \textbf{PSTNet+PointCMP} (50\% Semi-supervised)&  88.5 \\
    \bottomrule
    \end{tabular}
    \label{NTU}
\end{table}

\begin{table}[t]
    \centering
    \setlength{\abovecaptionskip}{0.1cm}
    \setlength{\belowcaptionskip}{-0.3cm}
    \small
    \caption{Action recognition accuracy (\%) of transfer learning on MSRAction-3D. Accuracy improvements against the supervised baseline are shown in subscript.}
    \setlength{\tabcolsep}{1mm}
    \begin{tabular}{l|l|cc}
    \toprule
    \multicolumn{1}{l|}{\multirow{2}{*}{\textbf{\makecell[l]{Methods}}}}
    & \multicolumn{1}{c|}{{\multirow{2}{*}{\textbf{Input}}}}
    & \multicolumn{2}{c}{\textbf{\#Frames}}\\
    \cline{3-4}
    & & 8 & 16   \\
    \midrule
    4D MinkNet\cite{minke}+ROP \cite{wang2021self} & Point+RGB & 86.31 & - \\
    MeteorNet~\cite{MeteorNet}+ROP \cite{wang2021self} & Point+RGB  & 85.40$_{\blue{\textbf{+4.26}}}$ & - \\
    \midrule
    \textbf{PSTNet + PointCMP}  & Point & 88.53$_{\blue{\textbf{+5.03}}}$ & 91.58$_{\blue{\textbf{+1.68}}}$ \\
    \bottomrule
    \end{tabular}
    \label{Transfer Learning}
\end{table}

\subsection{Transfer Learning}
To evaluate the generalization performance of our PointCMP, we conduct experiments by transferring pre-trained encoder to other datasets or tasks. 
Specifically, the encoder was first pre-trained on the cross-subject training set of NTU-RGBD, and then fine-tuned with an additional MLP head on MSRAction-3D, NvGesture, and SHREC’17. 

\textbf{Transfer to MSRAction-3D.} 
We first fine-tuned the pre-trained encoder on MSRAction-3D following the experimental settings in Sec.~\ref{Sec4.2}.
We compare our PointCMP with ROP~\cite{wang2021self} in Table~\ref{Transfer Learning}. Note that, since the official code for ROP is unavailable, we report its performance on 4D MinkNet~\cite{minke} and MeteorNet~\cite{MeteorNet} for comparison. 
Although PSTNet uses only points as input, our PointCMP facilitates this baseline to surpass ROP by over 2\% accuracy. In addition, our PointCMP introduces more significant accuracy improvements as compared to ROP (5.03\% vs. 4.26\%).

\textbf{Transfer to NvGesture and SHREC’17.} 
The encoder was further transferred from action recognition to gesture recognition through fine-tuning on NvGesture and SHREC’17. Specifically, the pre-trained model was fine-tuned for 100 epochs with a batch size of 16. The initial learning rate was set to 0.01 with a cosine decay strategy. During fine-tuning, 32 frames were utilized with 512/256 points sampled for each frame on NvGesture/SHREC’17.
We compare our fine-tuned models to previous supervised state-of-the-art methods in Table~\ref{NvGesture}. 
As we can see, after fine-tuning for 100 epochs, our PointCMP facilitates PSTNet to produce very competitive accuracy. In addition, our PointCMP also allows for faster convergence such that more significant improvements are achieved after fine-tuning for only 35 epochs (\eg, 78.9\% vs. 84.0\% on NvGesture).
This also shows the superior generalization capability cross different tasks of the representations learned by our PointCMP.

\begin{table}[t]
    \centering
    \setlength{\abovecaptionskip}{0.1cm}
    \setlength{\belowcaptionskip}{-0.3cm}
    \caption{Gesture recognition accuracy (\%) of transfer learning on NvGesture (NvG) and SHREC'17 (SHR).}
    \begin{tabular}{lcc}
    \toprule
    \textbf{Methods} &\textbf{NvG} &\textbf{SHR}\\
    \midrule
    FlickerNet~\cite{flickernet}           & 86.3    & -    \\
    PLSTM-base~\cite{min2020efficient}     & 85.9    & 87.6 \\
    PLSTM-early~\cite{min2020efficient}    & 87.9    & 93.5 \\
    PLSTM-PSS~\cite{min2020efficient}      & 87.3    & 93.1 \\
    PLSTM-middle~\cite{min2020efficient}   & 86.9    & 94.7 \\
    PLSTM-late~\cite{min2020efficient}     & 87.5    & 93.5 \\
    Kinet~\cite{zhong2022no}               & 89.1    & 95.2 \\
    \rowcolor{gray!20}
    PSTNet (35 Epochs)\cite{pstnet}        & 78.9    & 87.0 \\
    \rowcolor{gray!20}
    PSTNet (100 Epochs)\cite{pstnet}       & 88.4    & 92.1 \\
    \midrule
    \textbf{PSTNet + PointCMP} (35 Epochs)         &  84.0  & 90.8 \\
    \textbf{PSTNet + PointCMP} (100 Epochs) \ \ \  &  89.2  & 93.3 \\
    \bottomrule
    \end{tabular}
    \label{NvGesture}
\end{table}

\begin{figure*}[ht]
	\centering
    \setlength{\abovecaptionskip}{0.1cm}
    \setlength{\belowcaptionskip}{-0.4cm}
	\includegraphics[width=0.95\linewidth]{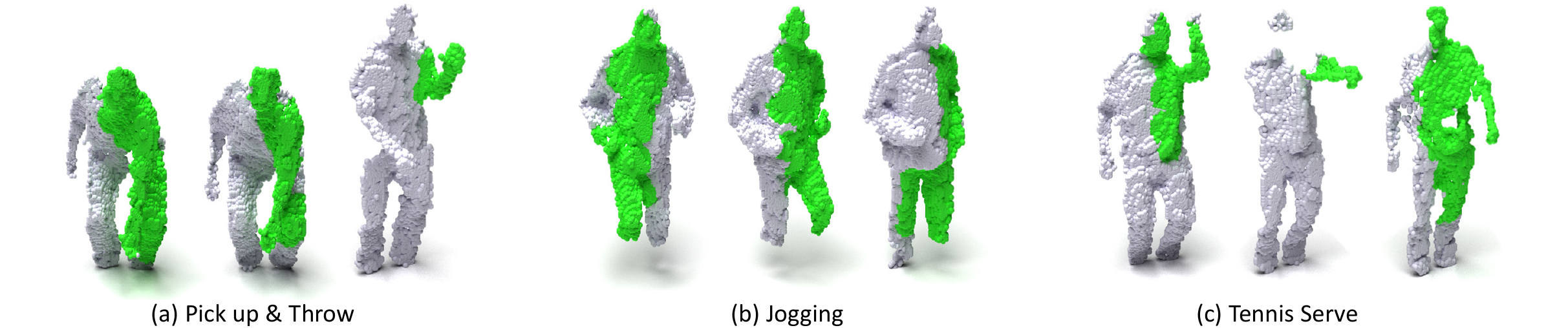}	\caption{Visualization of hard masked samples. Points corresponding to dominant tokens are marked in green.} 
	\label{fig6}
\end{figure*}

\subsection{Ablation Studies}
\label{Sec4.6}

\textbf{Architecture Design.} 
Our PointCMP employs a two-branch structure to simultaneously extract both local and global information, and adopts the mutual similarity based augmentation module to generate hard samples. To demonstrate the effectiveness of these architecture designs, we developed models A1 and A2 with only local and global branch, respectively. Then, model A3 is introduced by removing the mutual similarity based augmentation module. Quantitative results are presented in Table~\ref{architecture}.

As we can see, with only local or global branch, the performance of model A1 and A2 are limited (89.22\% and 49.49\%). This is because, both local and global information contribute to the recognition of point cloud videos. When these two branches are combined, complementary information can be exploited such that better accuracy is achieved by model A3 (89.76\%). However, without the mutual similarity based augmentation module, model A3 still suffers an accuracy drop of 2.16\% as compared to A4. This further validates the effectiveness of our mutual similarity based augmentation module.

\begin{table}[t]
    \centering
    \setlength{\abovecaptionskip}{0.05cm}
    \setlength{\tabcolsep}{1.1mm}
    \small
    \caption{Ablation studies on architecture designs.}
    \begin{tabular}{lccc|c}
    \toprule 
    & \textbf{\makecell[c]{Local\\Branch}} & 
    \textbf{\makecell[c]{Global\\Branch}} &  
    \textbf{\makecell[c]{Similarity-based\\Augmentation}} & 
    \textbf{Accuracy (\%)} \\
    \midrule
    \footnotesize{A1} &$\checkmark$   &                &              & 89.22 \\
    \footnotesize{A2} &               &  $\checkmark$  &              & 49.49 \\
    \footnotesize{A3} & $\checkmark$  &  $\checkmark$  &              & 89.76 \\
    \footnotesize{A4 (Ours)} & $\checkmark$  &  $\checkmark$  & $\checkmark$ & \textbf{91.92} \\
    \bottomrule
    \end{tabular}
    \label{architecture}
\end{table}

\begin{table}[t]
    \centering
    \setlength{\abovecaptionskip}{0.05cm}
    \setlength{\belowcaptionskip}{-0.3cm}
    \small
    \setlength{\tabcolsep}{0.85mm}
    \caption{Ablation studies on hard masked samples.}
    \begin{tabular}{ll|c|cccccc}
    \toprule
    \multicolumn{2}{r|}{\multirow{2}{*}{\textbf{Granularity}}} & \multirow{2}{*}{\textbf{Mask}} & \multicolumn{4}{c}{\textbf{Masking Ratio}} 	  \\
    \cline{4-7}
    &  &  & 25\%  & 50\%  & 75\%  & 90\%  	  \\
    \midrule
    \footnotesize{B1} & Token & Random           & 71.72 & 71.72 & 76.77 & 78.11 \\
    \footnotesize{B2} & Token & Similarity-based   & 70.03 & 81.82 & 84.18 & 88.55 \\
    \midrule
    \footnotesize{B3} & Segment & Random         & 90.81 & 88.15 & 79.80 & - 	  \\
    \footnotesize{B4 (Ours)} & Segment & Similarity-based & \textbf{91.92} & \textbf{90.24} & \textbf{86.53} & - 	  \\
    \bottomrule
    \end{tabular}
    \label{masking}
\end{table}

\textbf{Hard Masked Samples.} 
The masking strategy contributes to the quality of hard masked samples and plays a critical role in the local branch of our PointCMP. Consequently, we conduct experiments to study different masking strategies and compare their results in Table~\ref{masking}. 

As we can see, segment-wise masking strategy significantly outperforms token-wise masking strategy under different masking ratios. As compared to token-wise strategy, segment-wise strategy can better avoid the leakage of information caused by overlapped point patches, which facilitates the network to better exploit local structures in a point cloud video. 
Moreover, similarity-based masks introduce notable performance gains on segment-wise strategy, with accuracy being improved from 90.81\%/88.15\% to 91.92\%/90.24\%. This demonstrates the effectiveness of our hard masked samples. 

\begin{table}[t]
    \centering
    \setlength{\abovecaptionskip}{0.05cm}
    \small
    \caption{Ablation studies on hard negative samples.}
    \setlength{\tabcolsep}{1.7mm}
    \begin{tabular}{lc|c|c}
    \toprule 
    & \multicolumn{1}{r|}{\textbf{Hard Sample}} & \textbf{Strategy} & \textbf{Accuracy (\%)} \\
    \midrule
    \footnotesize{C1} & \ding{53}      &   -   &   90.52   \\
    \footnotesize{C2} & $\checkmark$ &   Random   &  91.29 \\
    \footnotesize{C3 (Ours)} & $\checkmark$ &   Similarity-based   &  \textbf{91.92}\\
    \bottomrule
    \end{tabular}
    \label{global hard negatives}
\end{table}

\begin{table}[t]
    \centering
    \small
	\setlength{\abovecaptionskip}{0.05cm}
    \setlength{\belowcaptionskip}{-0.4cm}
    \caption{{Ablation studies on the spatio-temporal matching module.}}
    \setlength{\tabcolsep}{1.2mm}
    \begin{tabular}{ll|c|c}
    \toprule 
    & \multicolumn{1}{l|}{\textbf{Architecture}} & \multicolumn{1}{l|}{\textbf{Matching Module}} & \textbf{Accuracy (\%)} \\
    \midrule
    \footnotesize{D1}        & Local         & \ding{53}     & 86.20 \\
    \footnotesize{D2}        & Local         &  $\checkmark$ & 89.22 \\
    \midrule
    \footnotesize{D3}        & Local \& Global & \ding{53}     & 90.24 \\
    \footnotesize{D4 (Ours)} & Local \& Global &  $\checkmark$ & \textbf{91.92} \\
    \bottomrule
    \end{tabular}
    \label{matching}
\end{table}

\begin{figure*}[t]
	\centering
    \setlength{\abovecaptionskip}{0.1cm}
    \setlength{\belowcaptionskip}{-0.4cm}
	\includegraphics[width=0.98\linewidth]{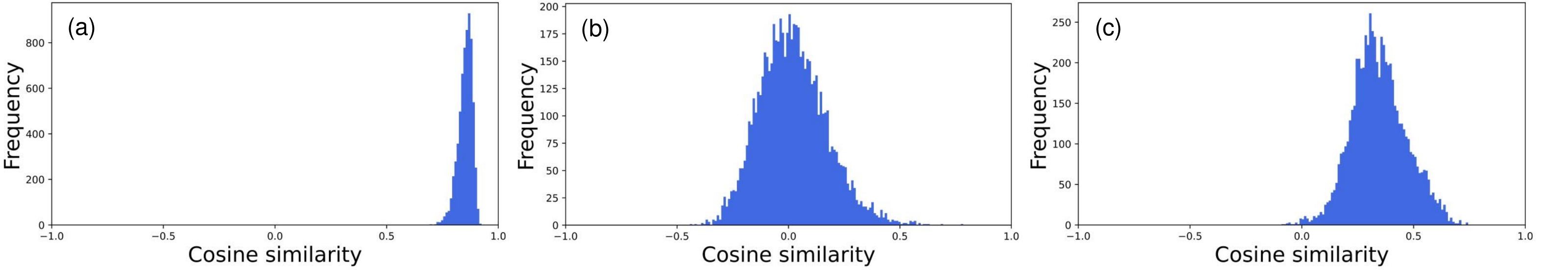}
	\caption{Visualization of cosine similarity histograms between the representations of query samples and their paired (a) positive samples, (b) negative samples, and (c) hard negative samples generated by channel erasing.}
	\label{fig5}
\end{figure*}

\begin{figure}[t]
    \centering
    \setlength{\abovecaptionskip}{0.1cm}
    \setlength{\belowcaptionskip}{-0.3cm}
    \includegraphics[width=0.95\linewidth]{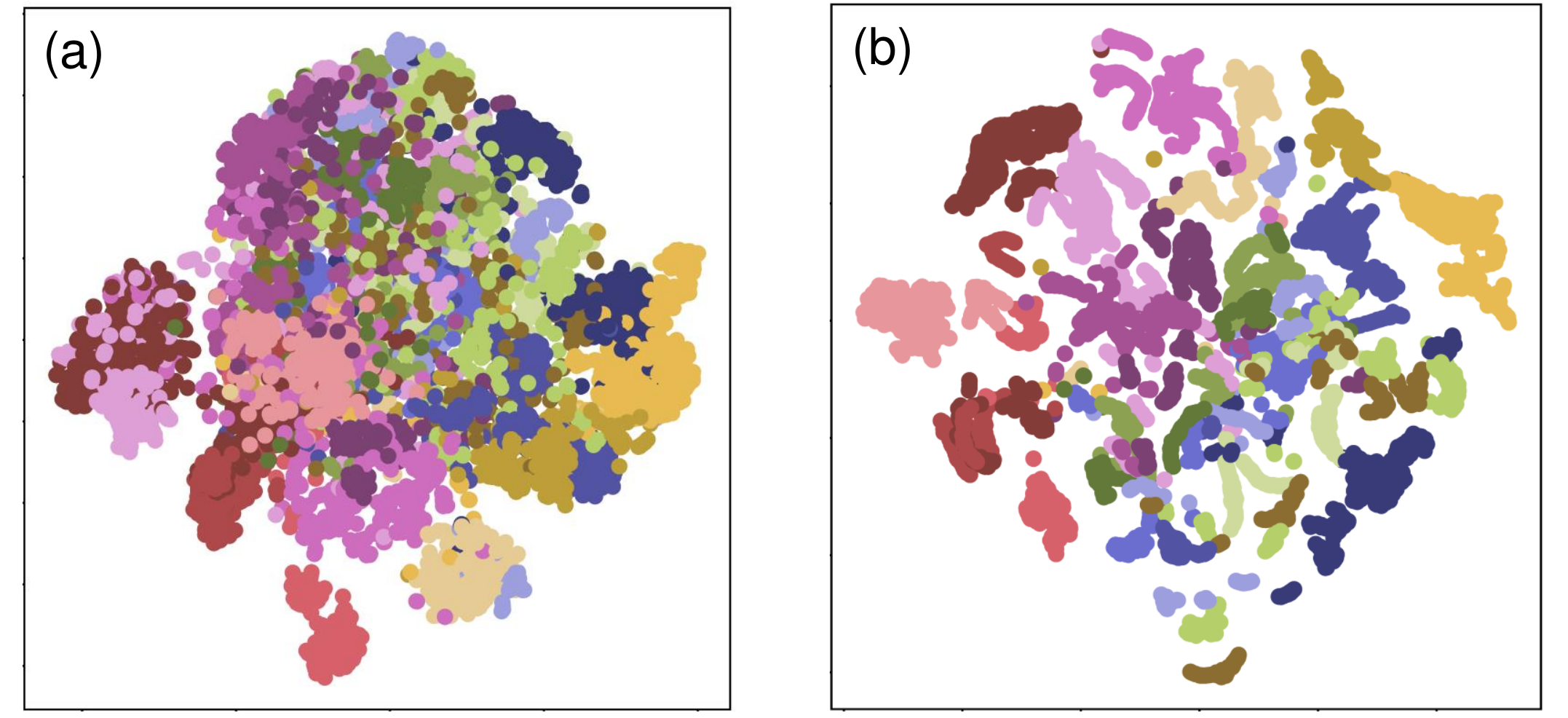}
    \caption{The t-SNE visualization of representation distributions on MSRAction-3D (a) after pre-training only with global contrastive learning and (b) after pre-training using our PointCMP.}
    \label{fig4}
\end{figure}

We further visualize the points corresponding to dominant tokens with high similarity to the global token in Fig.~\ref{fig6}. As we can see, tokens corresponding to moving body parts (e.g., arms in Fig.~\ref{fig6}(c)) are highlighted, which is consistent with our intuition. This demonstrates the feasibility of our mutual similarity based augmentation to synthesize reasonable hard samples.
With these discriminative regions being masked, the encoder is encouraged to leverage more context for mask prediction, with representations of higher quality being learned. 

\textbf{Hard Negative samples.} 
To demonstrate the effectiveness of hard negative samples in the global branch of our PointCMP, model C1 is introduced by excluding hard samples during training. That is, only samples in other videos are employed as negatives. Furthermore, we conduct experiments to study different channel erasing strategies. Quantitative results are presented in Table~\ref{global hard negatives}.

It can be observed that model C1 suffers an accuracy drop of 1.40\% as compared to C3 when hard negative samples are excluded. Using random channel erasing to generate hard negative samples, model C2 improves C1 with accuracy being increased from 90.52\% to 91.29\%. With our mutual similarity based augmentation module, hard negative samples of higher quality can be synthesized such that better performance can be achieved. This validates the effectiveness of the hard negatives generated by principal channel erasing. 

Following \cite{ge2021robust}, we visualize cosine similarity of representations learned from different sample pairs in Fig.~\ref{fig5} to study the importance of our hard negative samples. 
A model pre-trained on MSRAction-3D without using hard negatives is utilized for analyses. 
As shown in Fig.~\ref{fig5}(a), the similarities of positive pairs are close to 1 with an average value of 0.875.
On the contrary, negative pairs are gathered around 0 with an average value of 0.013. 
For our hard negatives, their average similarity score is increased to 0.315, which means these samples remain difficult for the pre-trained encoder if they are not included for training. This further shows the necessity of our hard negatives.

\begin{figure}[t]
	\centering
    \setlength{\abovecaptionskip}{0.1cm}
    \setlength{\belowcaptionskip}{-0.4cm}
	\includegraphics[width=0.96\linewidth]{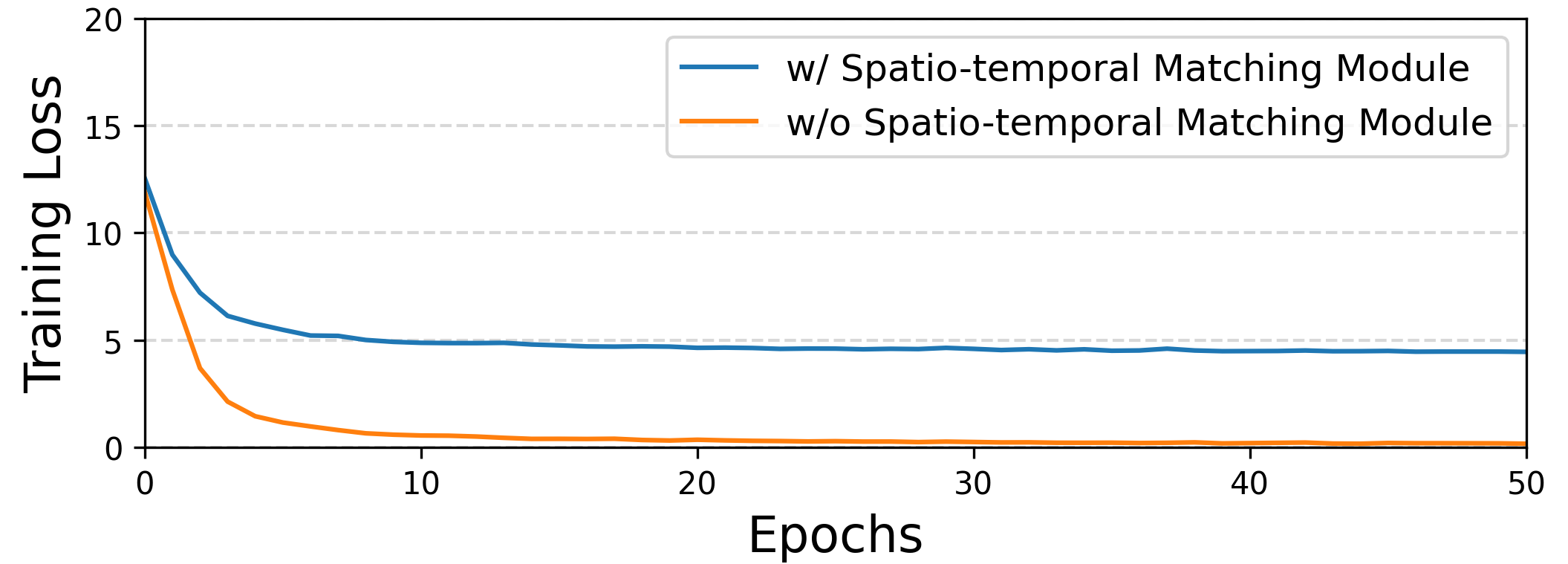}
	\caption{Evolution of local contrastive learning loss during pre-training on MSRAction-3D.}
	\label{fig777}
\end{figure}

\textbf{Spatio-temporal Matching Module.} In the local branch of our PointCMP, a spatio-temporal matching module is adopted to conduct local contrastive learning. {To study its effectiveness, we first developed model D2 with only the local branch. Then, we introduced model D1 and D3 by removing this matching module from D2 and D4, respectively. Quantitative results are presented in Table~\ref{matching}. As we can see, the spatio-temporal matching module facilitates D4 to produce an accuracy improvement of 1.68\% and introduces a more significant improvement of 3.02\% to D2.}
We further visualize the evolution of the local contrastive loss (i.e., $\mathcal{L}_{NCE}^{local}$ in Eq.~\ref{eq7}) in Fig.~\ref{fig777}. 
Without the spatio-temporal matching module, the loss decreases rapidly to near 0 and the networks cannot be further optimized. This is because the leakage of location information is leveraged by the network as shortcuts without capturing geometric information. In contrast, our matching module alleviates positional information leakage and increases the hardness of learning to help the network ultimately achieve higher accuracy (Table~\ref{matching}).

\textbf{Representation Visualization.} We further visualize the feature distributions using t-SNE to demonstrate the effectiveness of our PointCMP. 
With only global branch as many previous methods do, the learned representations have blurred boundaries between different categories with limited discriminative capability, as shown in Fig.~\ref{fig4}(a). 
In contrast, the representations extracted using our PointCMP can better exploit both global and local information with clearer boundaries between different categories, as shown in Fig.~\ref{fig4}(b). This clearly demonstrates the high discrimination of the representations learned by our method.

\section{Conclusion}
In this paper, we develop a self-supervised learning framework termed PointCMP for point cloud videos. Our PointCMP unifies the complementary advantages of contrastive learning and mask prediction paradigms to simultaneously learn both global and local spatio-temporal features at different granularities. To promote the training of PointCMP, we propose a mutual similarity based augmentation module to generate hard masked and negative samples at the feature level. Experiments on benchmark datasets show that our PointCMP achieves state-of-the-art performance on both action and gesture recognition tasks.

\noindent\textbf{Acknowledgments.} This work was partially supported by the National Natural Science Foundation of China (No. U20A20185, 61972435), the Guangdong Basic and Applied Basic Research Foundation (2022B1515020103), and Shanghai Science and Technology Innovation Action Plan (21DZ203700).

{\small
\bibliographystyle{ieee_fullname}
\bibliography{egbib}
}

\end{document}